\documentclass[conference]{IEEEtran}
\IEEEoverridecommandlockouts
\usepackage{cite}
\usepackage{amsmath,amssymb,amsfonts}
\usepackage{algorithmic}
\usepackage{graphicx}
\usepackage[linesnumbered,ruled,vlined]{algorithm2e}
\usepackage{textcomp}
\usepackage{xcolor}
\usepackage{multirow}
\def\BibTeX{{\rm B\kern-.05em{\sc i\kern-.025em b}\kern-.08em
    T\kern-.1667em\lower.7ex\hbox{E}\kern-.125emX}}
\begin{document}

\title{Video Compression with Hierarchical Temporal Neural Representation}

\author{Jun Zhu \qquad
    Xinfeng Zhang\qquad
    Lv Tang \qquad
    Junhao Jiang \qquad
    Gai Zhang\qquad
    Jia Wang\\
    University of Chinese Academy of Sciences\\
    {\tt\small \{zhujun23, xfzhang, jiangjunhao25, zhanggai16, wangjia242\}@mails.ucas.ac.cn,}\\
    {\tt\small luckybird1994@gmail.com}\\}

\maketitle

\begin{abstract}
Video compression has recently benefited from implicit neural representations (INRs), which model videos as continuous functions. INRs offer compact storage and flexible reconstruction, providing a promising alternative to traditional codecs. 
However, most existing INR-based methods treat the temporal dimension as an independent input, limiting their ability to capture complex temporal dependencies. To address this, we propose a Hierarchical Temporal Neural Representation for Videos, TeNeRV.
TeNeRV integrates short- and long-term dependencies through two key components. First, an Inter-Frame Feature Fusion (IFF) module aggregates features from adjacent frames, enforcing local temporal coherence and capturing fine-grained motion. Second, a GoP-Adaptive Modulation (GAM) mechanism partitions videos into Groups-of-Pictures and learns group-specific priors. The mechanism modulates network parameters, enabling adaptive representations across different GoPs.
Extensive experiments demonstrate that TeNeRV consistently outperforms existing INR-based methods in rate-distortion performance, validating the effectiveness of our proposed approach.
\end{abstract}

\begin{IEEEkeywords}
Video Compression, Implicit Neural Representation, Temporal Modeling
\end{IEEEkeywords}
\section{introduction}
With the advancement of modern multimedia, video compression has become increasingly important. 
Traditional hybrid video coding \cite{wien2015high,bross2021overview,wiegand2003overview} has evolved into a mature standard over decades. However, its modular design may lead to suboptimal synergy between components. In response, deep learning-based approaches \cite{lu2019dvc,li2021deep,li2023neural,li2024neural} have been integrated into video coding. These methods can extract spatiotemporal features learning from large datasets and achieve global optimality through end-to-end optimization. However, learning-based methods often suffer from limited generalization ability. For instance, the DCVC series codecs \cite{li2021deep,li2023neural,li2024neural} exhibit significant performance degradation when processing screen content videos \cite{tang2025canerv}, likely due to the absence of such data in training datasets.

To address this problem, methods such as NeRV \cite{chen2021nerv} utilize implicit neural representations (INRs) to model video sequences holistically. 
INR-based methods~\cite{mildenhall2021nerf, Tang_2023_ICCV} do not rely on a fixed prior learned from training data.
They treat videos as functions that map spatiotemporal coordinates to pixel values. 
A neural network is trained to approximate this mapping, with its parameters serving as a compact representation for video storage and reconstruction.
Their compression performance is primarily influenced by the model's representation capacity and inherent characteristics of the video.
HNeRV \cite{chen2023hnerv} improves coding efficiency by introducing an explicit feature storage structure, while HiNeRV \cite{kwan2024hinerv} further advances the approach with hierarchical encoding.

However, existing INR-based methods fail to match the performance of other compression approaches, particularly on highly dynamic videos due to the inefficient modeling of temporal information.
Unlike traditional codecs that leverage motion compensation to efficiently model inter-frame pixel trajectories, INRs rely on the holistic memorization of absolute pixel values across the temporal axis. 
In rapidly changing scenes, the visual discontinuity across frames imposes a heavy burden on shared network weights, leading to optimization interference among timestamps. 
Without explicit motion modeling to alleviate this temporal entropy, the limited network capacity is consumed by competing temporal patterns, resulting in over-smoothed artifacts and substantial loss of fine-details.

\begin{figure}[tp]
    \centering
    \includegraphics[width=\linewidth]{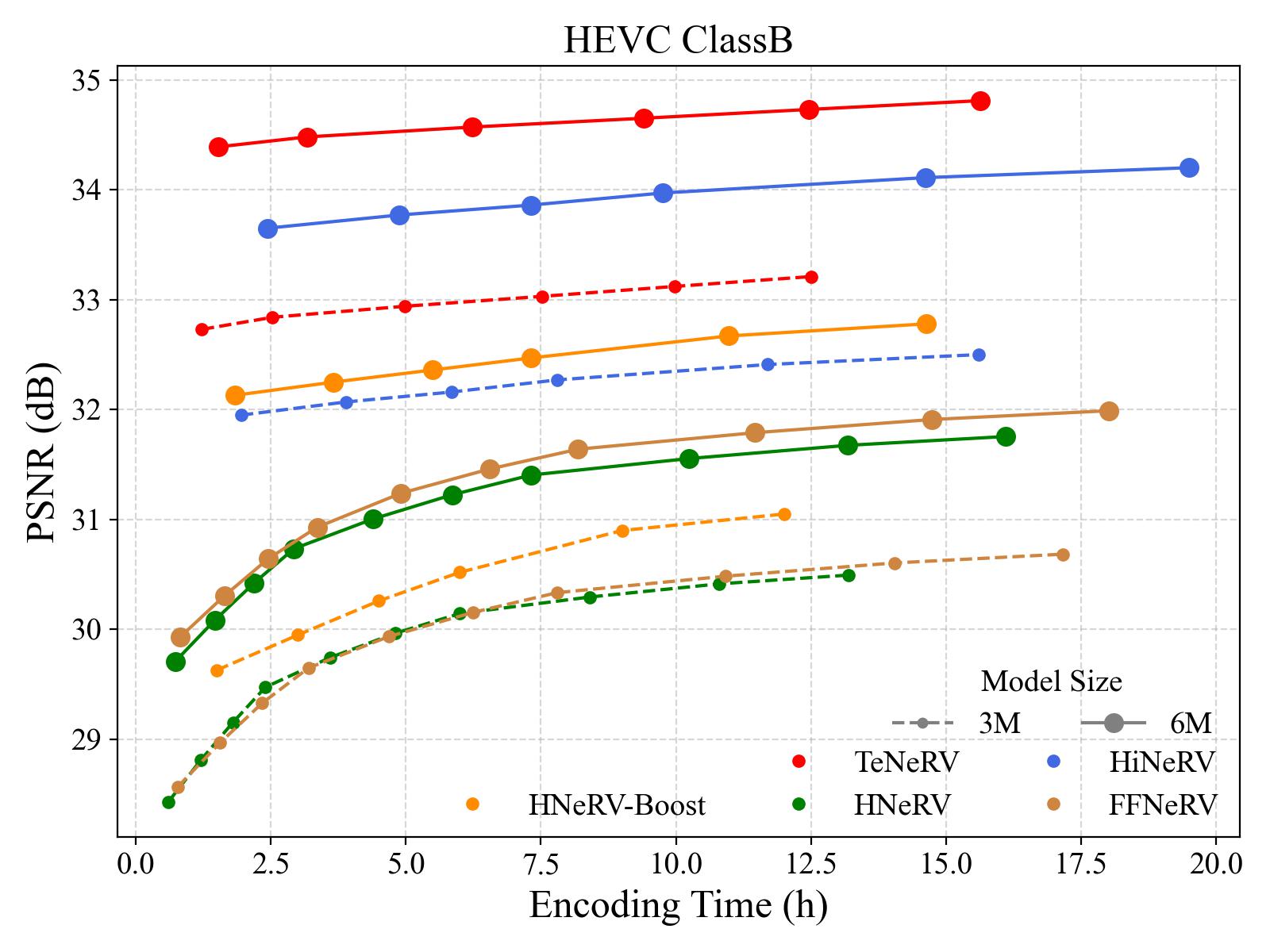}
    \caption{Reconstruction quality (PSNR) with different training times under 3M and 6M model size. TeNeRV(Ours) achieves the best performance.}
    \label{fig:psrn-time}
\end{figure}

\begin{figure*}[tp]
    \centering
    \includegraphics[width=\linewidth]{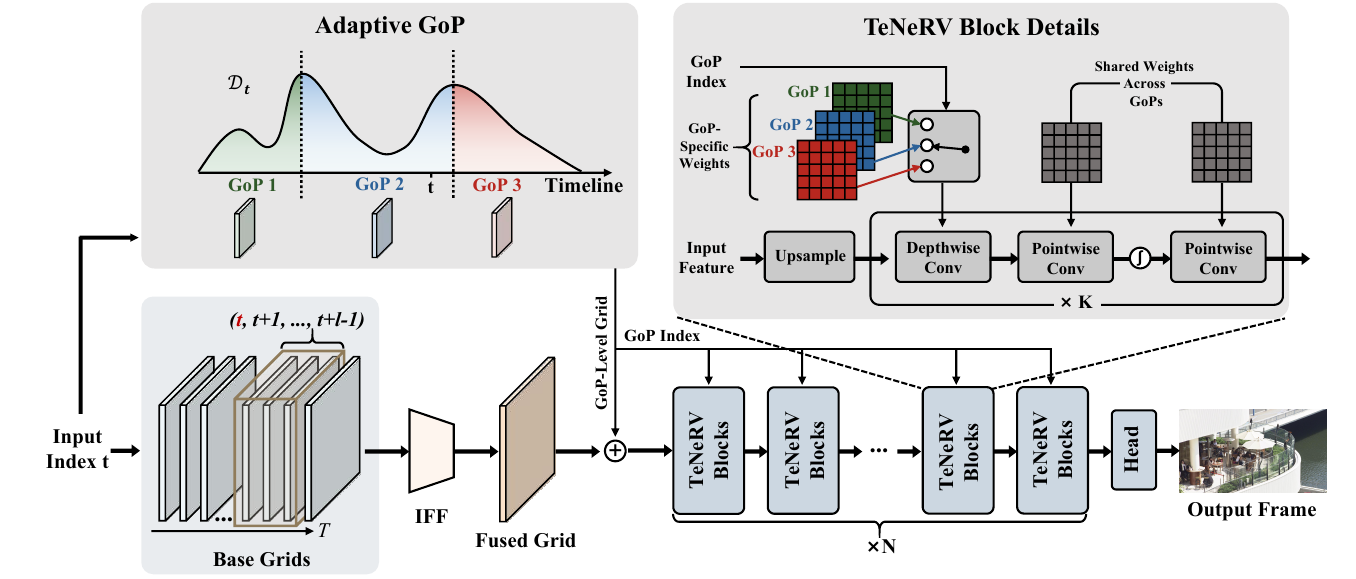}
    \caption{Overview of proposed TeNeRV architecture. The temporal fused grid is combined with the GoP-level grid and then upsampled through N TeNeRV blocks. The weights of these block are adaptive across different GoPs.}
    \label{fig:MTE}
\end{figure*}

To address this limitation, we propose a Hierarchical Temporal Neural Representation for Videos, TeNeRV. Unlike existing INR-based approaches that treat the temporal dimension $t$ merely as an independent input coordinate, 
TeNeRV explicitly decomposes the temporal dimension into a hierarchical structure to capture both global temporal consistency and local variations.
Specifically, we introduce a Inter-Frame Feature Fusion (IFF) network and a GoP-Adaptive Modulation (GAM) mechanism.
Instead of predicting frames in isolation, the IFF aggregates contextual information from adjacent timestamps to synthesize refined frame representations, thereby explicitly enforcing correlations between consecutive frames.
Based on these refined features, we further partition the video sequence into Groups-of-Pictures (GoPs) and assign each GoP a distinct learnable embedding. This GoP-level representation serves as a global anchor to capture group-specific priors. Simultaneously, the GAM mechanism also functions as a contextual condition to modulate the neural network, enabling the model to exhibit subtle, context-aware variations across different GoPs.
This framework effectively integrates short- and long-term dependencies, enabling precise capture of intricate motion and efficient bitrate allocation through optimized parameter distribution.
As shown in Fig.~\ref{fig:psrn-time}, when trained under the same time budget and with the same model capacity, TeNeRV achieves higher reconstruction performance than existing methods.

The main contributions of TeNeRV are as follows:
\begin{itemize}
    \item We design an Inter-Frame Feature Fusion module that aggregate features from adjacent timestamps. It explicitly enforces local temporal coherence and refines high-frequency motion details.
    \item We introduce a GoP-Adaptive Modulation mechanism that learns group-specific priors and dynamically modulate the network behavior across GoPs. This design allows the model to adapt flexibly to varying video content and scene transitions, improving the parameter efficiency.
    \item Extensive experiments demonstrate that TeNeRV achieves superior video compression performance compared with existing INR-based methods. Both quantitative and qualitative results validate the effectiveness of the proposed hierarchical temporal modeling strategy.
\end{itemize}


\section{Related work}

\subsection{Video Compression}
Video compression is a fundamental technology in compute vision and modern multimedia. 
Widely used standards such as H.264/AVC \cite{wiegand2003overview} and H.265/HEVC \cite{sullivan2012overview} are based on a hybrid framework that combines prediction, transformation, quantization and entropy coding. 
A key component of this framework is the Group of Pictures (GoP) structure, which organizes frames into a predefined temporal sequence. 
A GoP typically starts with an I-frame (Intra-coded frame), followed by a mix of P-frames (Predictive frames) and B-frames (Bidirectional frames).
Spatial and temporal predictions are employed in these frames for redundancy reduction.
Spatial prediction is primarily used in I-frames. It exploits the spatial correlation within a frame by predicting pixel values based on already encoded neighboring pixels. 
Temporal prediction is primarily used in P-frames and B-frames. It predicts entire blocks from previous or future frames.
The latest video compression standard H.266/VVC \cite{bross2021overview} achieves $\sim$50\% bitrate reduction compared to H.265/HEVC \cite{sullivan2012overview} at the same visual quality, by incorporating techniques like advanced block partitioning \cite{huang2021block}, and improved intra/inter predictions \cite{dong2021fast}.


\subsection{INR-based Video Compression}
Implicit Neural Representations (INRs) parameterize a signal as a continuous function that maps spatiotemporal coordinates $(x, y, t)$ or frame indices $t$ to pixel values. 
In the context of video compression, the paradigm shifts from storing pixel arrays to storing the parameters of the network trained to represent the video. 
Neural Representations for Videos (NeRV)\cite{chen2021nerv} pioneered this approach by formulating video compression as a frame-wise representation task. In NeRV, each frame index $t$ is mapped to a positional embedding, which is then decoded by an MLP-based generator into an RGB image. 
HNeRV~\cite{chen2023hnerv} improved upon NeRV by introducing a frame content-adaptive embedding as the network input.
FFNeRV~\cite{lee2023ffnerv} further incorporates flow information into frame-wise representations to exploit the temporal redundancy across the frames. The state-of-the-art, HiNeRV~\cite{kwan2024hinerv}, introduces hierarchical positional encoding. It enables the network to progressively refine visual information, leading to superior reconstruction quality.

\section{Method}
In this section, we first introduce the Inter-frame Feature Fusion (IFF) module (Section~\ref{subsec:CTF}). We describe the construction of base grids and the associated grid fusion mechanism.
Following this, we present the GoP-Adaptive Modulation (GAM) module (Section~\ref{subsec:GAM}), which includes the content-aware GoP partition strategy and the implementation of the modulation mechanism.
Finally, we present our training strategy (Section \ref{subsec:TS}), including the loss function and the overall training pipeline.
\subsection{Inter-frame Feature Fusion}
\label{subsec:CTF}
Existing INR-based methods employ various strategies to encode the temporal dimension $t$.
NeRV~\cite{chen2021nerv} adopts standard positional encoding (PE), mapping scalar time indices into high-frequency sinusoidal functions to distinguish different timestamps.
HNeRV~\cite{chen2023hnerv} utilizes a content-adaptive approach, employing a network to map frame indices directly to compact embeddings.
FFNeRV~\cite{lee2023ffnerv} and other grid-based methods\cite{kwan2024hinerv} utilize multi-resolution feature grids to capture temporal information at different scales.
Fundamentally, these methods share a common objective: mapping the low-dimensional temporal coordinate $t$ into a high-dimensional feature space. The semantic relationships between video frames will be implicitly reflected through the geometric distances in this spaces.
However, these approaches treat temporal encoding as a static mapping process, ignoring temporal variations or correlations between adjacent frames. Consequently, the network must infer motion dynamics from isolated codes, which often leads to temporal inconsistency and inefficient utilization of inter-frame redundancy.

In TeNeRV, we adopt the multi-resolution temporal grids proposed by FFNeRV as our base position embeddings:
\begin{equation}
    X_{base}^t = \gamma_{base}(t),0<t\leq T,
\end{equation}
where $X_{base}^i$ denotes the $i_{th}$ base grid, and $\gamma_{base}$ refers to a feature space composed of learnable grids.
Compared to  the Fourier encoding~\cite{chen2021nerv,chen2023hnerv,sitzmann2020implicit}, the base grid $X_{base}^t$ is better suited for video representation, providing a more flexible and content-adaptive temporal embedding.
Nonetheless, the structure alone cannot fully capture the temporal complexity present in real-world videos.
To address this, we employ a temporal window with learnable parameters $W_T=\{\overrightarrow{ w_1}, \overrightarrow{ w_2}, \dots, \overrightarrow{ w_T}\}$ to model diverse motions.
The base grids within window $t$ are fused using $\overrightarrow{ w_t}=\{w_t^1, w_t^2,\dots\}$.
Since the video content typically transitions gradually, adjacent frames have overlapping windows.
By applying a sliding window mechanism, we obtain a sequence of fused temporal grids, denoted as $\{X_{tem}^1,X_{tem}^2,\cdots,X_{tem}^T\}$.
The temporal grids with a window size $l$ can be written as 
\begin{equation}
    X_{tem}^t=\sum_{i=1}^{l}(w_t^i\cdot X_{base}^{t+i-1}),0<t\leq T .
\end{equation}
To ensure computational consistency at temporal boundaries, the base grid is expanded to size $T+l-1$. The base grid is then defined as follows: 
\begin{equation}
    X_{base}^i = \gamma_{base}(i),0<i\leq T+l-1.
\end{equation}
\subsection{GoP-Adpative Modulation}
\label{subsec:GAM}

\begin{algorithm}[t]
\caption{Content-Aware GoP Partitioning}
\label{alg:gop_partition_short}
\begin{algorithmic}[1]
\REQUIRE Divergence scores $\mathcal{D}$, target GoP count $K$, minimum length $L_{\min}$
\ENSURE GoP boundary set $\mathcal{B}$

\STATE $C \leftarrow \text{ArgsortDesc}(\mathcal{D})$\COMMENT{Candidate Boundaries}
\STATE $\mathcal{B} \leftarrow \{C_1, \dots, C_{K-1}\}$ 
\STATE $p \leftarrow K$\COMMENT{Pointer to next unused candidate}

\WHILE{true}
    \STATE $\mathcal{S} \leftarrow \text{Sort}(\mathcal{B} \cup \{0, T\})$
    \STATE Find $(u, v) \in \mathcal{S}$ such that $(v - u) < L_{\min}$
    
    \IF{no such pair exists}
        \RETURN $\mathcal{B}$
    \ENDIF

    \STATE $b \leftarrow \arg\min_{x \in \{u, v\} \cap \mathcal{B}} \mathcal{D}_x$
    \STATE $\mathcal{B} \leftarrow (\mathcal{B} \setminus \{b\}) \cup \{C_p\}$
    \STATE $p \leftarrow p + 1$
\ENDWHILE
\end{algorithmic}
\end{algorithm}

\begin{table*}[t]
\centering
\caption{
Video regression results (PSNR$\uparrow$) on HEVC ClassB \cite{sullivan2012overview} and UVG~\cite{mercat2020uvg}.
}
\begin{tabular}{c c c c c c c c| c c c c c c c c}
\noalign{\hrule height 1.2pt}

  & &\multicolumn{6}{c|}{HEVC ClassB} &\multicolumn{8}{c}{UVG}\\
 & Model & Bas. & BQT. & Cac. & Kim. & Par. & Avg. & Bea. & Bos. & Hon. & Joc. & Rea. & Sha. & Yac. & Avg.\\

\hline
\multirow{3}{*}{3M}
 & HNeRV &  27.16& 26.35& 28.85 &31.37 & 28.15& 28.43&33.40&34.74&39.20&30.96&24.55&34.59&28.52&32.28\\
 & FFNeRV  &     27.65& 26.43& 28.89 &31.33 & 28.06& 28.47&33.57&35.13&38.95&31.64&26.03&34.46&29.13&32.69\\ 
 & HNeRV-Boost & 28.49 & 27.79 & 30.09 & 32.55 & 29.71 & 29.63&33.70&35.52&39.50&33.62&27.60&35.50&29.10&33.51\\
 & HiNeRV       & 31.43 & 31.19 & 31.46 & 34.41 & 31.25 & 31.95&34.07&38.58&39.67&36.12&31.44&35.75&30.89&35.26\\
 & TeNeRV       & \textbf{32.48} & \textbf{31.87} & \textbf{32.30} & \textbf{35.40} & \textbf{32.35} & \textbf{32.84}&\textbf{34.17}&\textbf{38.80}&\textbf{39.71}&\textbf{37.22}&\textbf{32.34}&\textbf{35.85}&\textbf{31.30}&\textbf{35.63}\\
 
\hline
\multirow{3}{*}{6M}
 & HNeRV &28.66 & 27.73& 30.15 & 32.53 & 29.51& 29.71&33.90&36.19&39.49&33.07&27.23&35.81&30.07&33.68\\
 & FFNeRV      & 29.33 & 27.86 &30.39& 32.66 & 29.41& 29.93&33.96&36.35&39.52&33.09&27.26&35.83&30.09&33.73\\
 & HNeRV-Boost & 31.17 & 30.31 & 32.71 & 35.09 & 32.04 & 32.26&34.18&38.28&39.75&36.02&31.05&36.60&31.35&35.32\\
 & HiNeRV      & 33.45 & 32.54 & 33.14 & 36.41 & 32.86 & 33.68&34.30&40.14&39.79&37.72&34.30&36.75&32.71&36.53\\
 & TeNeRV      & \textbf{34.10} & \textbf{32.96} & \textbf{33.89} & \textbf{37.14} & \textbf{34.30} & \textbf{34.48}&\textbf{34.41}&\textbf{40.45}&\textbf{39.80}&\textbf{38.61}&\textbf{35.16}&\textbf{36.93}&\textbf{33.42}&\textbf{36.96}\\
 
\noalign{\hrule height 1.2pt}
\end{tabular}

\label{tab:regression}
\end{table*}

Over long temporal spans, video sequences often exhibit semantic shifts or scene transitions.
To handle such complex long-term dynamics, we introduce the Group-of-Pictures (GoP) concept, which decomposes the video into temporal segments.
Instead of a fixed-length GoP structure, we propose an adaptive partitioning algorithm guided by the content dynamics captured in the fused temporal grids. 
We identify scene transitions by tracking the feature drift between consecutive timestamps. Specifically, we define a divergence metric $\mathcal{D}_t$ based on the cosine distance between adjacent feature vectors:
\begin{equation}
    \mathcal{D}_t=1-\frac{X_{tem}^t\cdot X_{tem}^{t-1}}{||X_{tem}^t||_2||X_{tem}^t||_2}.
\end{equation}
As detailed in Algorithm~\ref{alg:gop_partition_short}, after computing the $\mathcal{D}_t$ for all timestamps in the video sequence, we select a set of timestamps $\mathcal{B}$ with the largest divergence numbers as initial GoP boundaries. Then, we iteratively remove boundaries that lead to GoPs shorter than the defined min length $L_{min}$.
These invalid boundaries are replaced in a greedy manner by the next best candidates ranked by $\mathcal{D}_t$, until all resulting GoPs satisfy the length constraint.

After segmenting the video sequence into $K$ variable-length GoPs, we assign a unique learnable grid to each GoP.
This GoP-level grid encodes temporally stable background features and is combined with the temporal grids: 
\begin{equation}
X_{fused}^{t}=X_{tem}^{t}+\gamma_{GoP}(k),0<t\leq T, 0<k\leq K,
\end{equation}
where $\gamma_{Gop}$ represents the feature space formed by GoP-level grids, and $k$ indexes the $k^{th}$ GoP containing the current time step $t$.
By capturing features shared within each local temporal segment, GoP-level grids enhance the network’s ability to model long-term consistency while preserving local dynamics.

As shown in Fig. \ref{fig:MTE}, the grid information corresponding to single frames, adjacent frames and GoP-level frames is integrated into the fused feature grids $X_{fused}^t$.
Subsequently, a cascade of TeNeRV Blocks is employed to progressively upsample the spatial resolution for final video reconstruction.
To accommodate the distinct visual distribution across different GoPs, the internal structure of these blocks is dynamically modulated based on the GoP index.

Each TeNeRV Block consists of an upsampling operator followed by a refinement module built upon depthwise separable convolutions. This architectural decouples spatial feature extraction (via Depthwise layers)  from channel-wise feature integration (via Pointwise layers).
Observing that spatial textures can vary significantly across GoPs while feature projection patterns remain relatively consistent, we employ a hybrid weight-sharing strategy. Specifically, GoP-specific weights are assigned to depthwise layers to capture scene-specific semantics, while the pointwise convolution weights are shared across the entire video sequence. This design achieves an effective balance between content adaptively and parameter efficiency.

\subsection{Training Strategy}
\label{subsec:TS}
In TeNeRV, we adopt a two stage training pipeline, including the holistic pretraining and the GoP-Adaptive training.
During the pretraining phase, the entire video sequence is treated as a unified signal without applying the GoP-dependent switching. The network is trained using a generic configuration where all weights are shared across timestamps. This stage produces a coarse  representation of the video content, which serves as a basis for subsequent GoP partitioning.

After the GoP structure is determined, the network is transitioned to its adaptive configuration for GoP-adaptive training.
The GoP-level grids and the GoP-specific depthwise convolution kernels are initialized by duplicating the corresponding weights learned during the pretraining stage.

\section{experiments}
\subsection{Set Up}
\subsubsection{Dataset}
To evaluate the effectiveness of TeNeRV, we conduct extensive experiments on UVG\cite{mercat2020uvg} and HEVC ClassB\cite{sullivan2012overview} datasets.
The UVG dataset contains seven video sequences with a resolution of 1920x1080. Similarly, the HEVC ClassB dataset includes five video sequences at 1920x1080 resolution. They cover a diverse range of scenes with varying motion complexity and texture details, providing a comprehensive assessment of our method's performance.
\subsubsection{Implementation Details}
We implement TeNeRV using the PyTorch framework. All training and evaluation are conducted on the NVIDIA RTX 3090 GPU.
During the training phase, we adopt the proposed two stage training pipeline with a total of 300 epochs using the Adam optimizer. For the first 30 epochs, we train the network with the L1 loss to establish the global structure. For the remaining 270 epochs, we employ a hybrid loss function combining MS-SSIM and L1 loss, aiming to balance pixel-level accuracy with structural similarity.

\subsection{Video Representation}
To evaluate the representation capcity of our proposed method, we compared TeNeRV against leading INR baselines, including HNeRV\cite{chen2023hnerv}, FFNeRV\cite{lee2023ffnerv}, HNeRV-Boost\cite{zhang2024boosting} and HiNeRV\cite{kwan2024hinerv}.
To ensure a fair comparison, we standardize the model sizes across all methods. Specifically, we assess reconstruction quality at two distinct parameter scales: 3M and 6M. Peak Signal-to-Noise Ratio (PSNR) is employed as the metric to quantify the pixel-wise fidelity of the reconstructed frames.

Tab.~\ref{tab:regression} presents the PSNR results of all methods. As demonstrated, TeNeRV achieves the best performance across different datasets and parameter scales.
Specifically, TeNeRV surpasses the state-of-the-art INR model, HiNeRV, by a substantial margin of over 0.8 dB on the HEVC ClassB and approximately 0.4dB on the UVG dataset on average.
Moreover, the performance gains are predominantly concentrated on sequences with more dynamic motion patterns (e.g. Jockey and Yacht) , which suggests the effectiveness of our method in temporal modeling.
Compared to other baselines, our method achieves a remarkable performance gain of more than 2.0 dB. This superior parameter efficiency validates the high representational capability of our proposed architecture.
\subsection{Time Complexity}

\begin{table}[t]
\centering
\caption{
Complexity comparison. All models are
evaluated on videos with 600 frames (BQTerrace) with 3M parameters. 
}
\begin{tabular}{c | c c }
\noalign{\hrule height 1.2pt}
 Method&Encoding time($\downarrow$) &  Decoding FPS ($\uparrow$)\\
 \hline
HNeRV &54min &142.9 img/s\\
FFNeRV  & 1h10min& 75.4 img/s\\
HNeRV-boost &1h30min& 63.9 img/s\\
HiNeRV     &2h13min& 54.3 img/s\\
TeNeRV &2h35min &48.1 img/s\\

\noalign{\hrule height 1.2pt}
\end{tabular}
\label{time}
\end{table}

\begin{table}[t]
\centering
\caption{
Abalation Study on HEVC ClassB. The positive values of BD-Rate reflect the bitrate increasing compared to TeNeRV.  
}
\renewcommand{\arraystretch}{1.15}
\begin{tabular}{l|cccccc}
\noalign{\hrule height 1.2pt}
Methods & TeNeRV & V1 & V2 & V3 & V4 & V5 \\
\noalign{\hrule height 0.8pt}
BD-rate (\%) & 0.00 & 5.1 & 4.9 & 10.3 & 3.4 & 2.7 \\
\noalign{\hrule height 1.2pt}
\end{tabular}

\label{tab:aba}
\end{table}

To evaluate the time complexity, we compared TeNeRV against other INR-based methods.
As shown in Tab. ~\ref{time}, when trained for a fixed schedule of 300 epochs, our method incurs slightly longer encoding and decoding times. However, per-epoch computational cost alone does not provide a complete assessment of efficiency. 
To obtain a more comprehensive evaluation, we train a series of independent models with different total numbers of training epochs in Fig.\ref{fig:psrn-time}.
As illustrated by these PSNR–Time curves,
our method consistently achieves higher reconstruction quality than competing approaches under identical training time and the same model scale.
Notably, the 3M variant of TeNeRV even outperforms most methods with 6M parameters.
These results suggest that our method maintains competitive, and even superior, time efficiency compared with existing methods.
Moreover, they also indicate that the performance gains of our approach do not stem from the increased computational cost, but rather from the inherent effectiveness of the hierarchical temporal modeling.

\begin{figure}[tp]
    \centering
    \includegraphics[width=\linewidth]{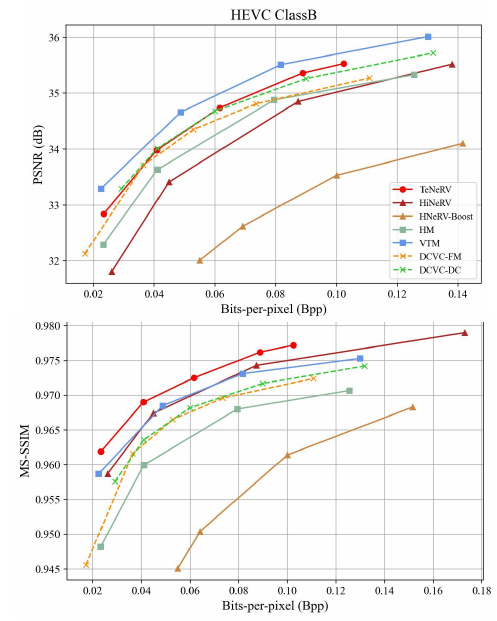}
    \caption{Video Compression results on the HEVC ClassB dataset.}
    \label{fig:RD}
\end{figure}
\subsection{Video Compression}
For video compression, we compare TeNeRV with traditional codecs ( H.266/VTM, H.265/HM), learning-based methods (DCVC-DC~\cite{li2023neural}, DCVC-FM~\cite{li2024neural}) and INR-based methods (HiNeRV, HNeRV-boost) on HEVC ClassB dataset.
To generate the bitstream for TeNeRV, we perform 30 epochs of quantization-aware training followed by entropy coding.
As shown in Fig.~\ref{fig:RD}, TeNeRV achieves superior rate–distortion (RD) performance compared with all neural video compression methods. Under the MS-SSIM metric, TeNeRV even outperforms the state-of-the-art codec VTM.

\begin{figure*}[ht]
\centering
\includegraphics[width=\linewidth]{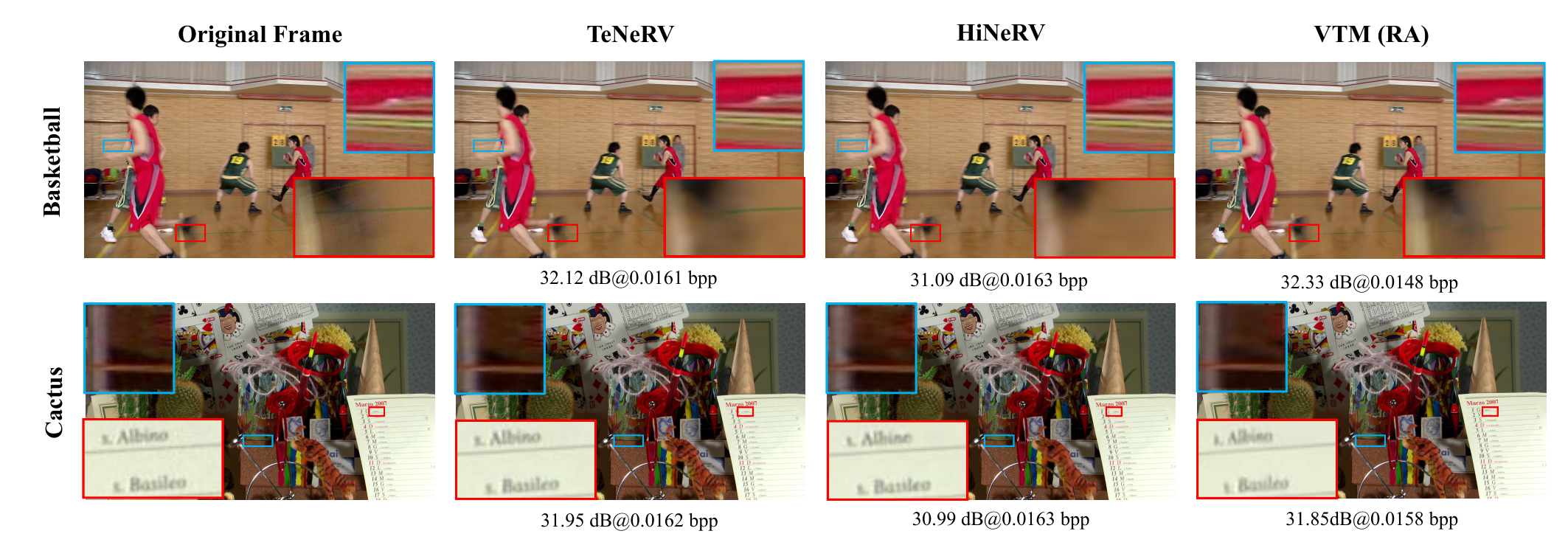} 
\caption{Visualization of video compression results on different videos. Reconstruction detail are shown in the rigion marked by red box. The motion area (blue box) is presented by stacking pixels at the same position across consecutive frames. }
\label{fig:visual_comparison}
\end{figure*}

Fig. \ref{fig:visual_comparison} presents qualitative comparisons on the BasketballDrive and Cactus datasets. 
The visualization results further illustrate that TeNeRV achieves more effective motion modeling compared to HiNeRV.
Moreover, both TeNeRV and HiNeRV outperform VTM in compression regions with structured patterns, such as the green line in the Basketball sequence and the text in Cactus. This is because such regions can be represented by relatively simple functions, making them more suitable for INR-based modeling.
For temporal comparisons, our method demonstrates improved temporal stability relative to HiNeRV and VTM, exhibiting smoother quality variations over time and avoiding pronounced temporal fluctuations.

\subsection{Ablation Studies}
We conduct ablation studies on the HEVC ClassB dateset to evaluate the effectiveness of our hierarchical temporal modules.
In the first variant (V1), we remove the IFF module and rely solely on base grids. As shown in Tab.~\ref{tab:aba}, this modification leads to a performance degradation of 5.1\%.
This decline can be directly attributed to the absence of explicit inter-frame correlation modeling, indicating that temporal dependencies between adjacent frames provide essential guidance for capturing local temporal dynamics.
In the second variant (V2), removing the GoP-level grids (V2) results in a 4.9\% performance drop. This observation suggests that the GoP-level priors serve as effective static anchors for modeling background content.
In the third variant (V3), we replace the GoP-specific depthwise kernels with a single shared weight.
This change causes a substantial performance degradation of 10.3\%, validating the necessity of the proposed content-adaptive weighting mechanism for handling scene variations.
Finally, we examine the impact of the GoP partitioning strategy. We compare our adaptive approach with fixed-length GoP schemes (V4) and heuristic strategies that depend solely on video length (V5). The results demonstrate that our adaptive partitioning algorithm consistently outperforms these content-agnostic baselines, highlighting the importance of dynamic temporal segmentation.

\section{Conclusion}
In this paper, we proposed TeNeRV, a hierarchical temporal neural representation for video compression that explicitly models both local temporal coherence and long-term semantic consistency. By integrating inter-frame feature fusion with adaptive GoP-level modulation, TeNeRV effectively captures complex motion patterns while maintaining high parameter efficiency. Extensive experiments demonstrate that TeNeRV consistently outperforms existing INR-based approaches in both rate–distortion performance and temporal stability under comparable model sizes and training budgets. These results indicate that our hierarchical temporal modeling is critical for advancing neural video compression.

\end{document}